# The Relationship Between Knowledge, Belief, and Certainty: Preliminary Report

Joseph Y. Halpern
IBM Almaden Research Center
San Jose, CA 95120
halpern@ibm.com (ARPA/CSNET)
halpern@almvma (BITNET)

**Abstract:** We consider the relation between knowledge and certainty, where a fact is *known* if it is true at all worlds an agent considers possible and is *certain* if it holds with probability 1. We identify certainty with probabilistic belief. We show that if we assume one fixed probability assignment, then the logic KD45, which has been identified as perhaps the most appropriate for belief, provides a complete axiomatization for reasoning about certainty. Just as an agent may believe a fact although $\varphi$ is false, he may be certain that a fact $\varphi$ is true although $\varphi$ is false. However, it is easy to see that an agent can have such false (probabilistic) beliefs only at a set of worlds of probability 0. If we restrict attention to structures where all worlds have positive probability, then S5 provides a complete axiomatization. If we consider a more general setting, where there might be a different probability assignment at each world, then by placing appropriate conditions on the *support of the probability function* (the set of worlds which have non-zero probability), we can capture many other well-known modal logics, such as T and S4. Finally, we consider which axioms characterize structures satisfying *Miller's principle*.

## 1 Introduction

A great deal of interest has focussed recently on logics of knowledge and probability (see, for example, the volumes [Hal86, Var88, KL86, KL87]). Researchers have used the possible-worlds approach to give semantics to knowledge by saying an agent *knows* a fact $\varphi$ if $\varphi$ is true at all the worlds the agent considers possible. We can also give semantics to formulas involving probability in a possible-worlds framework by saying $\varphi$ holds with probability $\alpha$ if the set of worlds where $\varphi$ is true is a set of probability $\alpha$.

It is well known that we can capture different notions of knowledge by varying the conditions on the *accessibility relation* which defines the set of worlds that an agent considers possible (see [HM85] for an overview). One particular set of requirements on the accessibility relation, namely that it be *serial, Euclidean, and transitive* (we define these terms below) results in the logic KD45 which has been considered the logic most appropriate for *belief* [Lev84a, FH88a].

We can also give a probabilistic interpretation to belief. The greater the probability of $\varphi$ (according to an agent's subjective probability function), the stronger an agent's belief is in $\varphi$. In this paper, we identify *certainty* – where an agent is said to be certain of $\varphi$ if $\varphi$ holds with probability 1 – with probabilistic belief. We show that such an identification is well motivated: If we have one fixed probability assignment on the set of possible worlds, then *certainty satisfies precisely the axioms of KD45*. In KD45, an agent may hold false beliefs; i.e., he may believe a fact $\varphi$ that is false. Similarly, an agent may be certain about a fact which is false. However, we show that an agent can have such false beliefs only at worlds with probability 0; i.e., almost surely, his (probabilistic) beliefs are correct. If we restrict attention to structures where all possible worlds have non-zero probability, then S5 gives a complete axiomatization for certainty: an agent no longer can have false beliefs.

We can extend these results by considering more general probability structures, where the agent may have a different probability function at each state of the world. Just as different axioms for knowledge can be captured by placing appropriate conditions on the set of worlds an agent considers possible, so different axioms for certainty can be captured by placing appropriate conditions on the *support* of the probability function, that is, the set of worlds to which the probability function assigns non-zero measure. Indeed, we show that many other well-known modal logics, such as T, D4, and S4, correspond in a natural way to conditions on the support.

This is not the first paper to consider the relationship between knowledge, belief, and certainty; Gaifman [Gai86] and Frisch and Haddawy [FH88c] also consider these issues. Both of these papers focus on structures that satisfy *Miller's principle* [Mil66, Sky80b] (this principle is discussed in detail later). Gaifman [Gai86] shows that the valid formulas of S5 are precisely those which hold with probability 1 in his logic (when restricted to structures satisfying Miller's principle), while Frisch and Haddawy [FH88c] argue that the valid formulas of the modal logic D4 are precisely those that hold with probability 1 in their logic (which is also intended to capture Miller's principle). We show that in our framework, there is a precise sense in which KD45 characterizes certainty in those structures satisfying Miller's principle. We remark that Morgan has also considered the relationship between axioms for probability and axioms for more standard modal logics [Mor82a, Mor82b], but his focus is on conditional probabilities and the results have a much different flavor from ours.

The rest of this paper is organized as follows. In the next section, we present the formal model for reasoning about probability (which is a slight variant of the model discussed in [FHM88, FH88b]). In Section 3 we review the formal semantics for reasoning about knowledge, stating a number of results that are needed in the sequel. In Section 4 we show that KD45, the logic of belief, is a complete axiomatization for reasoning about certainty (with respect to the probability structures introduced in Section 2), and that if we restrict attention to structures where all worlds have non-zero probability, then S5 is a complete axiomatization. In Section 5 we consider generalized probability structures, and show how different conditions on the support of the probability measure correspond to different axiomatizations. In particular, we show that many of the classical modal logics can be captured by placing the appropriate conditions on the probability structures. While these results are all quite straightforward, they do show an interesting and not altogether obvious connection between certainty and knowledge. In Section 6 we briefly discuss some extensions to our results. In Section 7, we consider structures satisfying Miller's principle and relate our results to those of [Gai86] and [FH88c]. We conclude in Section 8 with some further discussion.

## 2 Reasoning about probability

We are interested in making statements about certainty; that is we would like a logic that allows formulas of the form "The probability of $\varphi$ is 1." In order to accommodate such statements, we start with a more general logic, essentially that considered in [FHM88, FH88b]. In this logic, statements of the form $w(\varphi) \geq 1/2$ and $w(\varphi) < 2w(\psi)$ are allowed, which can be interpreted as "the probability of $\varphi$ is greater than or equal to $1/2$" and "the probability of $\varphi$ is less than twice the probability of $\psi$, respectively. More generally, linear combinations of expressions involving probability are allowed.

The formal syntax of the logic is quite straightforward. Well-formed formulas are formed by starting with primitive propositions, and closing off under Boolean connectives (conjunction and negation), as well as allowing *weight formulas* of the form $a_1 w(\varphi_1) + \ldots + a_k w(\varphi_k) \geq b$, where $a_1, \ldots, a_k, b$ are arbitrary integers and $\varphi_1, \ldots, \varphi_k$ are arbitrary formulas. We call the resulting language $\mathcal{L}^P$. A formula such as $w(\varphi) \geq 1/2$ is, strictly speaking, an abbreviation of the $\mathcal{L}^P$ formula $2w(\varphi) \geq 1$, while $w(\varphi) < 2w(\psi)$ is an abbreviation for $\neg(w(\varphi) \geq 2w(\psi))$. We also use a number of other obvious abbreviations without further comment, such as $w(\varphi) \leq b$ for $-w(\varphi) \geq -b$ and $w(\varphi) = b$ for $(w(\varphi) \geq b) \wedge (w(\varphi) \leq b)$.

Just as in [FH88b], we allow arbitrary nesting of probability formulas, so that $w(w(\varphi) \geq 1/2) < 1/3$ is a legal formula of $\mathcal{L}^P$.[1] Such higher-order probability statements will be one of our main interests here. They are not as unmotivated as they might first appear. Suppose we take $\varphi$ to be the statement "it will rain tomorrow," and we have just heard the weatherman say that it is likely to rain tomorrow. Thus, according to the weatherman, $w(\varphi) \geq 1/2$ holds. However, suppose we have found this weatherman to be quite unreliable in the past, so that his predictions turn out to be wrong far more often than they are right. Thus, we might place probability less than $1/3$ on his statement, which leads us exactly to the formula $w(w(\varphi) \geq 1/2) < 1/3$. (See [Gai86, Sky80b] for further discussion of higher-order probabilities.)

We use a possible-worlds approach to give semantics to the formulas in $\mathcal{L}^P$. (This is essentially the same approach as that taken by Nilsson [Nil86].) We take a *probability structure* $N$ to be a tuple $(S, \pi, pr)$, where $S$ is a finite or countably infinite set of *states* or *possible worlds*, $\pi$ associates with every state $s \in S$ a truth assignment $\pi(s)$ on the primitive propositions (so that $\pi(s)(p)$ is either **true** or **false** for every primitive proposition $p$ and state $s \in S$), and $pr$ is a discrete probability function on $S$ (so that $pr(s) \geq 0$ for each $s \in S$ and $\sum_{s \in S} pr(s) = 1$). We can think of $pr$ as being the agent's subjective probability assignment to the worlds in $S$. As usual, for every subset $A \subseteq S$, we define $pr(A) = \sum_{s \in A} pr(s)$. We have restricted $S$ here to be countable and $pr$ to be a discrete probability function for ease of exposition. We discuss in Section 6 how our results can be extended to more general settings.

---

[1] Nesting was not considered in [FHM88] to simplify the presentation, although there is no technical difficulty involved in adding it there as well.

We can now define the satisfaction relation $\models$, where $(N,s) \models \varphi$ is read "$\varphi$ is *true* (or *satisfied*), in state $s$ of the probability structure $N$", by induction on the structure of $\varphi$. The definitions for the propositional connectives are the standard ones. Intuitively, we would like a formula such as $w(\varphi) \geq 1/2$ to be true if the probability of the set of states where $\varphi$ is true is at least $1/2$. To make this precise, given a formula $\psi$, suppose we have defined $(N,s) \models \psi$ for all states $s \in S$. Let $S_\psi = \{s \in S : (N,s) \models \psi\}$. Then we define $(N,s) \models w(\varphi) \geq 1/2$ if $pr(S_\varphi) \geq 1/2$. The complete formal definition of $\models$ is given below:

$(N,s) \models p$ (for a primitive proposition $p$) iff $\pi(s)(p) = $ **true**

$(N,s) \models \varphi \wedge \psi$ iff $(N,s) \models \varphi$ and $(N,s) \models \psi$

$(N,s) \models \neg\varphi$ iff $(N,s) \not\models \varphi$

$(N,s) \models a_1 w(\varphi_1) + \cdots + a_k w(\varphi_k) \geq b$ iff $a_1 pr(S_{\varphi_1}) + \cdots + a_k pr(S_{\varphi_k}) \geq b$.

As usual, we say a formula $\varphi$ is *valid with respect to probability structure* $N = (S, \pi, pr)$, written $N \models \varphi$, if $(N,s) \models \varphi$ for all $s \in S$. A formula is *valid with respect to a class $\mathcal{N}$ of probability structures*, written $\mathcal{N} \models \varphi$, if $N \models \varphi$ for all $N \in \mathcal{N}$. Similarly, we say $\varphi$ is *satisfiable with respect to* $N$ if $(N,s) \models \varphi$ for some $s \in S$, and satisfiable with respect to $\mathcal{N}$ if $\varphi$ is satisfiable with respect to $N$ for some $N \in \mathcal{N}$.

In [FHM88], a complete axiomatization is provided for the sublanguage of $\mathcal{L}^P$ that allows only Boolean combinations of weight formulas with propositional arguments (i.e., if $\varphi$ occurs in the context $w(\varphi)$, then $\varphi$ is a propositional formula), while in [FH88b], techniques are sketched for extending this axiomatization to the full logic (indeed, in [FH88b], a complete axiomatization is provided for a richer language with modal operators for knowledge). Here, our interest is in a different sublanguage of $\mathcal{L}^P$, where the only probability statements are those that involve certainty, that is, those of the form $w(\varphi) = 1$ (with nesting allowed); we abbreviate such a formula as $Cert(\varphi)$. We call this sublanguage $\mathcal{L}^C$. Thus, a typical formula of $\mathcal{L}^C$ is $\neg q \wedge Cert(\neg Cert(p) \wedge Cert(q))$.

## 3 Reasoning about knowledge

The possible-worlds model can also be used to capture reasoning about knowledge. We briefly review the necessary ideas here; the interested reader is referred to [HM85] for more details.

The intuitive idea is that an agent knows $\varphi$ if $\varphi$ is true in all the worlds the agent considers possible. For now, we restrict our discussion to a situation involving only one agent; in Section 6 we discuss how our results can be extended to a situation involving many agents.

In order to reason about knowledge, we use a modal logic with a modal operator $K$, where $K\varphi$ is read "the agent knows $\varphi$." The well-formed formulas are formed by starting with primitive propositions, and closing off under Boolean connectives and applications of $K$. Thus, if $\varphi$ and $\psi$ are formulas, then so are $\neg\varphi$, $\varphi \wedge \psi$, and $K\varphi$. We call this language $\mathcal{L}^K$.

In order to give semantics to such formulas, we use *knowledge structures*.[2] A knowledge structure $M$ is a tuple $(S, \pi, \mathcal{K})$, where $S$ is a set of states (not necessarily countable), $\pi$ associates a truth assignment with every state in $S$, just as in the case of probability structures, and $\mathcal{K}$ is a binary relation on $S$. Intuitively, $(s,t) \in \mathcal{K}$ if, in state $s$, the agent considers $t$ possible. For future reference, we define $\mathcal{K}(s) = \{t : (s,t) \in \mathcal{K}\}$; thus, $\mathcal{K}(s)$ is the set of states the agent considers possible in state $s$.

Again, we define truth for formulas in $\mathcal{L}^K$ by induction on structure. The only clause that differs from that for $\mathcal{L}^C$ is that for formulas of the form $K\varphi$:

$(M,s) \models K\varphi$ iff $(M,t) \models \varphi$ for all $t$ such that $(s,t) \in \mathcal{K}$.

This captures the intuition that the agent knows $\varphi$ in state $s$ if $\varphi$ is true at all the worlds that the agent considers possible in state $s$.

We define validity and satisfiability with respect to a knowledge structure and a class of knowledge structures just as in the case of probability structures.

We are often interested in classes of knowledge structures where certain restrictions are placed on

---

[2] Our usage of the term *knowledge structure* here differs from that of [FHV84]. We use it here in contrast to probability structures.

144

the binary relation $\mathcal{K}$, since by restricting $\mathcal{K}$ we can capture a number of interesting properties of knowledge. Recall that a binary relation $\mathcal{K}$ on $S$ is *reflexive* if $(s,s) \in \mathcal{K}$ for all $s \in S$, *transitive* if $(s,t) \in \mathcal{K}$ and $(t,u) \in \mathcal{K}$ implies $(s,u) \in \mathcal{K}$, *symmetric* if $(s,t) \in \mathcal{K}$ implies $(t,s) \in \mathcal{K}$, *Euclidean* if $(s,t) \in \mathcal{K}$ and $(s,u) \in \mathcal{K}$ implies $(t,u) \in \mathcal{K}$, and *serial* if for all $s \in S$, there is some $t$ such that $(s,t) \in \mathcal{K}$. Let $\mathcal{M}$ be the class of all knowledge structures. We restrict $\mathcal{M}$ by using superscripts $r$, $s$, $t$, $e$, and $l$ to denote reflexive, symmetric, transitive, Euclidean, and serial structures, respectively. Thus, $\mathcal{M}^{rt}$ denotes the class of all reflexive and transitive knowledge structures, $\mathcal{M}^{elt}$ denotes the class of Euclidean, serial, and transitive structures, and so on.

Consider the following collection of axioms:

**P** All instances of axioms of propositional logic

**K** $(K\varphi \land K(\varphi \Rightarrow \psi)) \Rightarrow K\psi$

**T** $K\varphi \Rightarrow \varphi$

**4** $K\varphi \Rightarrow KK\varphi$

**5** $\neg K\varphi \Rightarrow K\neg K\varphi$

**D** $\neg K\mathit{false}$

and rules of inference:

**R1** From $\varphi$ and $\varphi \Rightarrow \psi$ infer $\psi$

**R2** From $\varphi$ infer $K\varphi$[3]

We get various systems by combining some subset of **K**, **T**, **4**, **5**, and **D** with **P**, **R1**, and **R2**. Thus, we get the logic K by combining **K** with **P**, **R1**, and **R2**, KT by combining **K** and **T** with **P**, **R1**, and **R2**, and so on. Traditionally, KT4 has been called S4, and KT45 has been called S5; KD45 is sometimes called weak S5 [FH88a]. As well, the K is often omitted, so that KT becomes T, KD becomes D, and so on. We try to use the most common notation throughout this paper, and hope the reader will bear with us.

Different authors have argued for the appropriateness of different logics to capture knowledge. For example, S5 has been used to capture a notion of knowledge appropriate for analyzing distributed systems [HM84, Hal87] and synchronous digital machines [RK86]. Moore used S4 in [Moo85]. On the other hand, since the knowledge represented in a knowledge base is typically not required to be true, T has been thought to be inappropriate for these applications; thus, KD45 is considered, for example, in [Lev84a]. KD45 is also considered to be an appropriate logic for characterizing the beliefs of an agent who might believe things that in fact turn out to be false [FH88a, Lev84b].

We say that an axiom system $A$ is *sound* with respect to a class of (knowledge or probability) structures $\mathcal{Q}$ if all the axioms in $A$ are valid with respect to $\mathcal{Q}$ and the rules of inference preserve validity; $A$ is *complete* with respect to a class $\mathcal{Q}$ if all the valid formulas in $\mathcal{Q}$ are provable using the axioms and rules of inference of $A$.

It turns out that there is a close connection between conditions placed on $\mathcal{K}$ and the axioms. In particular, **T** corresponds to $\mathcal{K}$ being reflexive, **4** to $\mathcal{K}$ being transitive, **5** to $\mathcal{K}$ being Euclidean, and **D** to $\mathcal{K}$ being serial. To make this precise, we define an axiom system $A$ to be *normal* if it consists of the axioms **P**, **K**, rules of inference **R1**, **R2**, and some subset (possibly empty) of the axioms **T**, **4**, **5**, and **D**. The class of structures *corresponding to* $A$ is that class that results by restricting to the relations corresponding to the axioms as discussed above. For example, $\mathcal{M}^{elt}$ is the class corresponding to KD45 and $\mathcal{M}^r$ is the class corresponding to T. We use $\mathcal{M}^A$ to denote the class of structures corresponding to the normal axiom system $A$. We then get the following well-known result (whose proof can be found in [Che80, HM85]):

**Theorem 3.1:** *If $A$ is a normal axiom system, then $A$ is sound and complete with respect to $\mathcal{M}^A$ (for the language $\mathcal{L}^K$).*

As a consequence of Theorem 3.1, we get, for example, that KD45 is a sound and complete axiomatization with respect to $\mathcal{M}^{elt}$ and that T is a sound and complete axiomatization with respect to $\mathcal{M}^r$. Since a binary relation is reflexive, symmetric, transitive (i.e., an equivalence relation) iff and only if it is reflexive, Euclidean, and transitive, we get that S5 is a sound and complete axiomatization with respect to $\mathcal{M}^{rst}$.

We need two more results from modal logic. The proof of the first can be found in [Che80, HM85]. It says that although we have allowed the set of states in a knowledge structure to be infinite and even uncountable, we can without loss of generality (at least as far as satisfiability and validity are concerned) restrict attention to finite knowledge structures, i.e., those where the set of states is finite. We say a formula $\varphi$ is *consistent* with an axiom system $A$ if $\neg\varphi$ cannot be proved from $A$.

---

[3] The names **K**, **T**, **4**, **5**, and **D** are fairly standard, and are taken from [Che80]. The axiom **D** given in [Che80] is different from that given here, although the two versions are equivalent in the presence of **P**, **K**, **R1**, and **R2**.

**Theorem 3.2:** *If $A$ is a normal axiom system and $\varphi$ is consistent with $A$, then $\varphi$ is satisfiable in a finite knowledge structure in $\mathcal{M}^A$.*

The second result relates S5 provability to KD45 provability. The result is undoubtedly known to experts, although a proof does not seem to appear in the literature.

**Theorem 3.3:** *The formula $\varphi$ is S5 provable iff $K\varphi$ is KD45 provable.*

## 4 Relating certainty and knowledge

We first show that if we consider $\mathcal{N}_0$, the class of all probability structures as defined in Section 2, then certainty is characterized by the axioms of KD45. We first define some notation: if $\varphi$ is a formula in $\mathcal{L}^C$, let $\varphi^K$ be the formula in $\mathcal{L}^K$ that results by replacing all occurrences of $Cert$ by $K$. Similarly, if $\varphi$ is a formula in $\mathcal{L}^K$, let $\varphi^C$ be the formula in $\mathcal{L}^C$ that results by replacing all occurrences of $K$ by $Cert$. For each axiom system $A$ discussed in Section 3, let $A^C$ be the result of replacing all occurrences of $K$ in the axioms and inference rules of $A$ by $Cert$.

**Theorem 4.1:** $KD45^C$ *is a sound and complete axiomatization for the language $\mathcal{L}^C$ with respect to $\mathcal{N}_0$.*

**Corollary 4.2:** *If $\varphi$ is a formula in $\mathcal{L}^K$, then $\varphi$ is S5 provable iff $\mathcal{N}_0 \models Cert(\varphi^C)$.*

Corollary 4.2 is closely related to Theorem 5 of [Gai86]; we discuss the precise relationship in Section 7.

Note that KD45 allows the agent to have false beliefs; $\neg\varphi \wedge K\varphi$ is consistent with KD45. By interpreting $K$ as certainty (by translating a formula $\varphi$ to $\varphi^C$), we get some added insight into the probability of having false beliefs. Given a probability structure $N = (S, \pi, pr)$, let $FB$ consist of those states $s \in S$ where the agent has some false beliefs, i.e., those states $s$ where for some formula $\varphi$ we have $(N, s) \models \neg\varphi \wedge Cert(\varphi)$. Then it is easy to see that $FB$ is a set of measure 0.

**Proposition 4.3:** $pr(FB) = 0$.

Proposition 4.3 shows that if there are no states of measure 0, then the agent will not have false beliefs. This suggests that S5 will form a complete axiomatization in this case. To make this precise, let $\mathcal{N}_1$ consist of those probability structures where all states have positive measure (thus $N = (S, \pi, pr) \in \mathcal{N}_1$ iff $pr(s) > 0$ for all $s \in S$).

**Theorem 4.4:** $S5^C$ *is a sound and complete axiomatization with respect to $\mathcal{N}_1$.*

It is well-known that using the axioms of KD45, we can prove that any formula in $\mathcal{L}^K$ is equivalent to a formula with no nesting of $K$'s. (This is proved by using the equivalences $K(\varphi \wedge \psi) \equiv (K\varphi \wedge K\psi)$, $KK\varphi \equiv K\varphi$, $K\neg K\varphi \equiv K\varphi$, $K(\varphi \vee K\psi) \equiv (K\varphi \vee K\psi)$, and $K(\varphi \vee \neg K\psi) \equiv (K\varphi \vee \neg K\psi)$, all of which are easily seen to be valid with respect to $\mathcal{M}^{elt}$; we omit details here.) Using Theorem 4.1, it follows that

**Corollary 4.5:** *For every formula $\varphi$ in $\mathcal{L}^C$, there is a formula $\varphi'$ which has no nesting of $Cert$ such that $\varphi$ is equivalent to $\varphi'$ in all probability structures; i.e., $\mathcal{N} \models \varphi \equiv \varphi'$.*

*A fortiori*, the result also holds for $\mathcal{N}_1$. This says that we do not gain any expressive power by allowing nesting of the $Cert$ modality. Note that we do gain expressive power if we can make statements that involve probabilities other than 1; the formula $w(w(p) \geq 1/2) < 1/3$ is not equivalent to any formula without nested probability statements.

## 5 Generalized probability structures

There are situations for which the probability structures discussed in Section 2 may not be general enough to capture what is going on. In particular, since there is only one probability function in the picture, we cannot capture situations where there is some uncertainty about the probability function.

For example, consider an agent tossing a coin, which he knows to be either a fair coin (so that the probability of both heads and tails is 1/2) or a biased coin (so that the probability of heads is, say, 1/3, while the probability of tails is 2/3). This suggests that we allow one possible world where the probability function assigns probability 1/2 to the event heads (i.e., to the set of possible worlds where the coin lands heads) and another possible world where the probability function assigns probability 1/3 to the event heads. We might even consider a situation where the agent does not know his own probability function (this is analogous to situations regarding the modelling of knowledge, where we want to allow an agent who does not know what he knows), and thus considers a number of worlds possible where he has different probability functions.

These scenarios lead us to a more general approach: associating a (possibly different) probability function with each possible world. We capture this intuition by means of *generalized probability structures*. A generalized probability structure $N$ is a tuple $(S, \pi, PR)$, where $S$ is a finite or countably infinite set of states, $\pi(s)$ is a truth assignment to the primitive propositions for each state $s \in S$, and $PR(s)$ is a probability function on $S$ for each state $s \in S$. Generalized probability structures can be viewed as a generalization of knowledge structures. Instead of just having a set of states that an agent considers possible from each state $s$, each world that the agent considers possible is assigned a probability (where the worlds that the agent does not consider possible are assigned probability 0). We remark that the *Kripke structures for knowledge and probability* of [FH88b] are in fact a generalization of generalized probability structures (in that they allow many agents and include modal operators for knowledge).

We give semantics to probability formulas just as before, except that when evaluating the truth of a weight formula in the state $s$, we use the probability function $PR(s)$. Thus, we get

$$(N, s) \models a_1 w(\varphi_1) + \ldots a_k w(\varphi_k) \geq b \quad (1)$$

$$\text{iff } a_1 PR(s)(S_{\varphi_1}) + \ldots a_k PR(s)(S_{\varphi_k}) \geq b. \quad (2)$$

Note that the probability structures of Section 2 can be viewed as a special case of generalized probability structures, where $PR(s) = pr$ for all states $s \in S$.

When reasoning about certainty, it is clear that, in some sense, all that is relevant are the states with non-zero measure. Given a generalized probability structure $N = (S, \pi, PR)$, let the *support* relation $Supp_N$ on $S$ be defined by: $(s, t) \in Supp_N$ if $PR(s)(t) > 0$; i.e., $(s, t) \in Supp_N$ if the probability function in state $s$ assigns positive probability to state $t$. It is easy to check from the definitions that $(N, s) \models Cert(\varphi)$ iff $(N, t) \models \varphi$ for all $t$ such that $(s, t) \in Supp_N$. This suggests that the $Supp_N$ relation plays the same role in generalized probability structures as the $\mathcal{K}$ relation does in knowledge structures. To make this precise, given a generalized probability structure $N = (S, \pi, PR)$, let $M_N = (S, \pi, \mathcal{K})$ be the knowledge structure where $\mathcal{K} = Supp_N$. Then we have

**Theorem 5.1:** *If $\varphi \in \mathcal{L}^C$, then $(N, s) \models \varphi$ iff $(M_N, s) \models \varphi^K$.*

For reasoning about knowledge, we obtain different axioms by varying the conditions on the relation $\mathcal{K}$. We can obtain analogous axioms for reasoning about certainty by varying the conditions on the support relation. Note that the support relation is always serial: there must be at least one state $t$ such that $PR(s)(t) > 0$, since if we sum $PR(s)(t)$ over all states $t$ we get 1. We can impose other restrictions on the support relation, just as we did for the accessibility relation $\mathcal{K}$; we then get analogous classes of generalized probability structures $\mathcal{N}^r$, $\mathcal{N}^{elt}$, and so on. Just as in the case of knowledge, given a normal axiom system $A$, we can talk about the class of generalized probability structures $\mathcal{N}^A$ corresponding to $A$.

**Theorem 5.2:** *If $A$ is a normal axiom system that includes $\mathbf{T}$ or $\mathbf{D}$, then $A^C$ is sound and complete with respect to $\mathcal{N}^A$ (for the language $\mathcal{L}^C$).*

This result shows that most of the standard logics of knowledge can be interpreted as logics of certainty.

## 6 Extensions

As we mentioned above, we can easily extend our structures to allow for many agents. Suppose we have many agents, each with his own subjective probability function. In the case of probability structures, this would amount to considering structures of the form $(S, \pi, pr_1, \ldots, pr_n)$, where $pr_i$ is agent $i$'s probability function, while in the case of generalized probability structures, we would have structures of the form $(S, \pi, PR_1, \ldots, PR_n)$, where $PR_i(s)$ is agent $i$'s probability function in state $s$. We would then extend the language to allow formulas of the form $Cert_i(\varphi)$: agent $i$ is certain that $\varphi$ holds. The analogous changes can be made in the case of knowledge structures too, and the axiom systems can be extended in the obvious way to allow reasoning about many agents (cf. [HM85]). All our results then go through with essentially no change.

Gaifman's HOPs (higher-order probability structures) are equivalent to generalized probability structures with two agents, one of which is taken to be the agent doing the reasoning, and the other which is taken to be the expert. The agent's probability function is taken to be independent of the state (and so is like the probability function in our probability structures in Section 2), while the expert may have a different probability function at each state. It is not quite clear why the expert has different probability functions in each state while the agent does not, but in any case Gaifman's model can be easily extended to allow the agent to have different probability functions at each state. Gaifman goes on to consider *general*

*HOPs*, in which the expert's probability function can be time-dependent. We can easily deal with this in our framework by adding temporal operators, and a temporal accessibility relation.

Frisch and Haddawy [FH88c] present a structure along the same lines as those of Gaifman, except that they actually allow the agent to have different probability functions at each state. However [Had], they view their structure as only appropriate for giving semantics to formulas with depth of nesting at most two (and thus inappropriate for a formula of the form $Cert(Cert(Certp)))$. In order to deal with deeper nesting, they require a whole sequence of probability functions. This makes their approach for nested formulas quite different from ours and that of Gaifman.

Another way we can extend our structures is by dropping the assumptions that the set of possible worlds is countable and that the probability function is discrete. We briefly discuss how to do so here.

If we drop the assumption that the probability function is discrete, we have to explicitly describe with each probability function its domain, the set of sets to which the function assigns a probability. These sets are called the *measurable sets*. We then have to slightly redefine the semantics of $Cert(\varphi)$ to take into account the possibility that the set $S_\varphi$ might not be measurable. If $N$ is a probability structure, we define $(N,s) \models Cert(\varphi)$ if there is some measurable set $A$ such that $A \subseteq S_\varphi$ and $\mu(A) = 1$. This essentially amounts to considering the *inner measure* induced by $\mu$ (see [FHM88, FH88b] for more details). It is easy to check that this definition agrees with our old definition if $S_\varphi$ is measurable. We make similar modifications if $N$ is a generalized probability structure. In this case, we also redefine the support relation so that $(s,t) \in Supp_N$ iff $t \in \cap_{\{A:PR(s)(A)=1\}} A$. Again, this definition agrees with our old definition of support if all sets are measurable. We leave it to the reader to check that, with these modifications, all our proofs go through with essentially no change. These modifications also enable us to deal with the case that the set of possible worlds is uncountable. We leave details to the reader.

## 7 Miller's principle

Gaifman [Gai86] and Frisch and Haddawy [FH88c] are mainly interested in structures that embody Miller's principle [Mil66, Sky80a, Sky80b]. In [Sky80a, Sky80b], a number of variants of Miller's principle are presented. The one of most interest to us here can be expressed

$$w_1(\varphi|(w_2(\varphi) \in I)) \in I,$$

where $w_1$ and $w_2$ can be viewed as the probability functions of two agents (we present possible interpretations for these agents below) and $I$ is an interval, which for the purposes of this discussion we can take to be a closed interval $[a,b]$ where $a$ and $b$ are rational endpoints with $0 \leq a \leq b \leq 1$. Intuitively this says that the conditional probability of $\varphi$ with respect to $w_1$, given that the probability of $\varphi$ with respect to $w_2$ is $a$, is $a$.

There are a number of possible interpretations of $w_1$ and $w_2$. One is to view $w_1$ as referring to *rational degrees of belief* of an agent and $w_2$ to refer to *propensities* or *objective probabilities* (see [Sky80a, Sky80b, Hal89] for further discussion of these issues). Another viewpoint is taken by Gaifman [Gai86], where as we mentioned in the previous section, $w_1$ is taken to represent the expert and $w_2$ the agent about whom we are reasoning. A third possibility mentioned by Skyrms [Sky80a] is that we can identify $w_1$ and $w_2$ as degrees of belief of an agent who does not necessarily know his own mind. This last interpretation is easily captured within a generalized probability structure. We focus on that interpretation for now, and then relate our results to those of Gaifman and Frisch and Haddawy.

Since we assume that $w_1$ and $w_2$ in Miller's principle now represent the same probability function, we replace both by $w$. This still does not does not correspond to a formula in $\mathcal{L}^P$, since we do not allow conditional probabilities. But since $w(\varphi|\psi) = w(\varphi \wedge \psi)/w(\psi)$, Miller's principle can be rewritten as

$$aw(w(\varphi) \in I) \leq w(\varphi \wedge (w(\varphi) \in I)) \leq bw(w(\varphi) \in I), \quad (*)$$

where we take $I$ to be the interval $[a,b]$ and $w(\varphi) \in I$ to be an abbreviation of $a \leq w(\varphi) \leq b$. This is (an abbreviation of) a formula in $\mathcal{L}^P$.[4]

Miller's principle (the axiom $(*)$) for the full language $\mathcal{L}^P$ is not sound with respect to any of the classes of structures we have considered so far. This is perhaps not surprising, since information about support is not sufficient to capture an axiom that talks about arbitrary probabilities, rather than just certainty. In [FH88b], probability structures satisfying a condition called *uniformity* are considered;

---

[4] Note that our requirement that $I$ be an interval with *rational* endpoints is necessary in order to make this a formula in $\mathcal{L}^P$. We also remark that rather than expressing the conditional probability as one term divided by another, we have cleared the denominator to avoid having to deal with the problems that arise when the denominator is 0.

these arise naturally in distributed systems applications. In the notation of this paper, a generalized probability structure $N = (S, \pi, PR)$ is *uniform* if for all $s, t \in S$, if $(s, t) \in Supp_N$, then $PR(s) = PR(t)$. As we now show, uniform structures do capture Miller's principle.

To make this precise, define a *probability frame* to be a pair $F = (S, PR)$, where $S$ is a set of states and $PR(s)$ is a discrete probability function on $S$ for each $s \in S$. Thus, a frame is a (generalized) probability structure without the truth assignment $\pi$. A probability structure $(S', \pi', PR')$ is *based on* frame $(S, PR)$ if $S = S'$ and $PR = PR'$. Uniformity and all the conditions on support that we have considered can be viewed as conditions on frames, rather than conditions on structures, since they do not depend on the truth assignment at all. Thus, for example, we can define a frame $(S, PR)$ to be uniform if for all $s, t \in S$, if $(s, t) \in Supp_F$, then $PR(s) = PR(t)$. Note a frame $F$ is uniform iff some probability structure based on $F$ is uniform iff every probability structure based on $F$ is uniform. We say a formula $\varphi$ is *valid in frame* $F$, written $F \models \varphi$, if $N \models \varphi$ for every probability structure $N$ based on $F$. The following theorem shows that Miller's principle characterizes uniform frames.

**Theorem 7.1:** *The following two conditions are equivalent:*

1. *$F$ is a uniform frame.*
2. *Every instance of Miller's principle (i.e., the axiom (*)) is valid in $F$.*[5]

We now want to show that when we restrict to reasoning about certainty, KD45 provides a complete axiomatization for uniform structures. Let $\mathcal{N}^{unif}$ be the class of uniform structures.

**Theorem 7.2:** *If $\varphi$ is a formula in $\mathcal{L}^C$, then $\mathcal{N}^{elt} \models \varphi$ iff $\mathcal{N}^{unif} \models \varphi$.*

**Corollary 7.3:** *$KD45^C$ is a sound and complete with respect to $\mathcal{N}^{unif}$ for the language $\mathcal{L}^C$.*

Thus $KD45^C$ is a sound and complete axiomatization for the class of structures characterized by Miller's principle, given our interpretation of Miller's principle with $w_1$ and $w_2$ identified. Combining Corollary 7.3 with Theorem 3.3, we get

**Corollary 7.4:** *If $\varphi$ is a formula in $\mathcal{L}^K$, then $\varphi$ is provable in S5 iff $\mathcal{N}^{unif} \models Cert(\varphi^C)$.*

As we mentioned above, Gaifman does not identify $w_1$ and $w_2$; rather, he associates $w_1$ with the expert's probabilities and $w_2$ with the agent's probabilities. In addition, Gaifman actually considers a stronger form of Miller's principle, namely

$$w_1(\varphi|\psi \land (w_2(\varphi) \in I)) \in I,$$

where $\psi$ is a conjunction of formulas of the form $w_2(\psi' \in I')$.[6] He shows that structures satisfying this stronger principle can be characterized in a way rather similar to Theorem 7.1. We provide a few of the details here. Suppose we are given a generalized probability structure $N = (S, \pi, PR_1, PR_2)$ for two agents. Recall that Gaifman assumes that the agent's probability assignment $PR_2$ is independent of the state, thus there is a fixed probability function $pr$ such that $PR_2(s) = pr$ for all states $s \in S$. Given a state $s \in S$, let $C(s)$ be the equivalence class of states in $S$ consisting of all states $s'$ such that $PR_1(s') = PR_1(s)$. Thus, $C(s)$ is the set of states where the expert has the same probability function that he does in $s$. Now consider the set of states $S_{good}$ where the expert believes that with probability 1, his probability distribution is the right one, i.e., $S_{good} = \{s : PR_1(s)(C(s)) = 1\}$. Gaifman shows that the stronger form of Miller's principle is equivalent to the condition $pr(S_{good}) = 1$, which Frisch and Haddawy call the *equivalence class constraint*. Namely, Gaifman shows that the stronger form of Miller's principle is sound in all structures satisfying the

---

[5] We remark that using frames to characterize axioms is a well-known technique in modal logic [Gol87, HC84]. Consider, for example, the axiom **T** for knowledge. Although we have noted that it is sound for *reflexive* knowledge structures (i.e., knowledge structures where the $\mathcal{K}$ relation is reflexive) and, together with **P**, **K**, **R1**, and **R2** provides a complete axiomatization for such structures, it is not hard to construct a non-reflexive structure where every instance of **T** is valid (see [HM85]). On the other hand, **T** does characterize reflexive *knowledge frames* (where a knowledge frame is just a pair $(S, \mathcal{K})$, and a reflexive knowledge frame is a knowledge frame where the relation $\mathcal{K}$ is reflexive), in that every instance of **T** is valid in a knowledge frame $F$ iff $F$ is reflexive. Similar remarks hold for all the other axioms we considered for knowledge. However, although Miller's principle does characterize uniform frames, it is *not* the case that Miller's principle together with the other axioms of probability discussed in [FHM88, FH88b] provides a complete axiomatization for the language $\mathcal{L}^P$ with respect to uniform frames. For example, the formula $(w(p) > a) \Rightarrow w(w(p) > a) = 1$, which is valid in uniform frames, can be shown not to be provable from Miller's principle and the other axioms. (Roughly speaking, this is because a model where probabilities get values in a *non-standard* field can be found where all of the axioms of probability and Miller's principle hold, and this formula is not satisfied.)

[6] It is not hard to extend our proof that Miller's principle is sound in uniform structures to show that if we identify $w_1$ and $w_2$, then this stronger principle is sound in uniform structures as well.

149

equivalence class constraint, and that the equivalence class constraint holds in all frames where the stronger form of Miller's principle is valid. Moreover, Gaifman shows that if $\varphi$ is provable in S5, then $\varphi$ holds with probability 1 (with respect to $pr$) in all structures satisfying the equivalence class constraint. Note that by Corollary 7.4, the analogous result holds for uniform structures.

As mentioned in the previous section, in [FH88c], where they only discuss probability formulas of depth 2, Frisch and Haddawy do not assume that $PR_2$ is the same for all agents. Since they also want to consider structures where Miller's principle holds (they do not consider the stronger version of Miller's principle), they assume that the equivalence class constraint holds for every probability function $PR_2(s)$, i.e., they assume that $PR_2(s)(S_{good}) = 1$ for all states $s$. However, in order to deal with more deeply nested formulas, Frisch and Haddawy plan to have a sequence of probability functions. Thus, in their structures they will have functions $PR_1, PR_2, PR_3, \ldots$, where $PR_i(s)$ gives a probability function at each state $s$. They do not provide interpretations for these probability functions, but they want to consider structures where Miller's principle holds between all consecutive pairs of probability functions given by $PR_i$ and $PR_{i+1}$. Thus, they plan to assume that the equivalence class constraint holds between consecutive pairs of probability functions. Miller's principle is easily seen to be sound in structures satisfying this constraint. Rather than having a different modal operator for each of these probability assignments, they only have one modal operator $Cert$. They use the $PR_i$'s to give semantics to more deeply nested occurrences of $Cert$. More formally (using our notation), in order to interpret an occurrence of the modal operator $Cert$ appearing at depth $i$ at state $s$, they use the probability function $Prob_i(s)$, where $Prob_1(s) = PR_1(s)$ and for $i > 1$, we define $Prob_i(s)$ as the mean value of $PR_i(s')$ over all states $s'$, where the weighting is done with respect to $Prob_{i-1}$. Thus, if $S'$ is a subset of $S$, then $Prob_i(s)(S')$ is $\sum_{s' \in S} PR_i(s)(s') \cdot Prob_{i-1}(s')(S)$ [Had]. They show that under this interpretation for $Cert$, the axioms of $KD4^C$ are sound, but not necessarily complete.

## 8 Conclusions

We have examined the relationship between knowledge, belief, and probability. We showed that, just as we can capture different properties of knowledge by placing appropriate conditions on the accessibility relationship, we can capture properties of certainty by placing conditions on the support. In particular, if we assume one fixed probability function, we showed that KD45 provides a complete axiomatization. Moreover, the set of worlds where an agent has false beliefs has probability 0. Interestingly, KD45 also provides a complete axiomatization with respect to structures satisfying Miller's principle.

Many researchers have rejected S5 as an appropriate axiomatization for an agent's beliefs since they want to allow an agent to have false beliefs. Instead, they consider KD45, which allows false beliefs. Our results suggest that there is a reasonable interpretation for belief that is characterized by the KD45 axioms, but still makes rather strong assumptions about the correctness of an agent's beliefs.

These results show how the tools of modal logic can be brought to bear on reasoning about probability. We believe that further work along these lines should yield further insights into probability, belief, and knowledge.

**Acknowledgments:** This paper was inspired by discussions with Peter Haddawy. Moshe Vardi made a number of useful comments on an earlier draft of the paper.